\documentclass[preprint,10pt]{elsarticle}

\usepackage{pifont}
\usepackage{xcolor}
\newcommand{\cmark}{\textcolor{green!80!black}{\ding{51}}}
\newcommand{\xmark}{\textcolor{red}{\ding{55}}}

\usepackage{svg}
\usepackage{amsmath}
\usepackage{tikz}
\usepackage{graphicx}
\usepackage{caption}
\usepackage{comment}

\usepackage{amsfonts}
\usepackage{mathtools}
\usepackage{framed}
\usepackage{microtype}
\usepackage{adjustbox}
\usepackage{epsfig}
\usepackage{url}

\usepackage{lipsum}

\usepackage{adjustbox}
\usepackage{floatrow}
\usepackage{graphics}

\usepackage[font=small,labelfont=bf]{caption}

\usepackage{graphicx}
\usepackage{xcolor}
\usepackage{array}
\usepackage{amsmath}
\usepackage{amsfonts}
\usepackage{soul}

\usepackage{amssymb}
\journal{}

\begin{document}

\begin{frontmatter}

%% Title, authors and addresses

%% use the tnoteref command within \title for footnotes;
%% use the tnotetext command for theassociated footnote;
%% use the fnref command within \author or \address for footnotes;
%% use the fntext command for theassociated footnote;
%% use the corref command within \author for corresponding author footnotes;
%% use the cortext command for theassociated footnote;
%% use the ead command for the email address,
%% and the form \ead[url] for the home page:
%% \title{Title\tnoteref{label1}}
%% \tnotetext[label1]{}
%% \author{Name\corref{cor1}\fnref{label2}}
%% \ead{email address}
%% \ead[url]{home page}
%% \fntext[label2]{}
%% \cortext[cor1]{}
%% \affiliation{organization={},
%%             addressline={},
%%             city={},
%%             postcode={},
%%             state={},
%%             country={}}
%% \fntext[label3]{}

\title{L-MAE: Longitudinal masked auto-encoder with time and severity-aware encoding for diabetic retinopathy progression prediction}

%% use optional labels to link authors explicitly to addresses:
%% \author[label1,label2]{}
%% \affiliation[label1]{organization={},
%%             addressline={},
%%             city={},
%%             postcode={},
%%             state={},
%%             country={}}
%%
%% \affiliation[label2]{organization={},
%%             addressline={},
%%             city={},
%%             postcode={},
%%             state={},
%%             country={}}

\affiliation[inst1]{organization={LaTIM UMR 1101, Inserm},%Department and Organization
            state={Brest},
            country={France}}

\affiliation[inst2]{organization={University of Western Brittany},%Department and Organization
            state={Brest},
            country={France}}

\affiliation[inst3]{organization={IMT Atlantique},%Department and Organization
            state={Brest},
            country={France}}

\affiliation[inst4]{organization={Ophtalmology Department, University Hospital of Brest},%Department and Organization
            state={Brest},
            country={France}}

\affiliation[inst5]{organization={Lariboisière Hospital, AP-HP},%Department and Organization
            state={Paris},
            country={France}}

\affiliation[inst6]{organization={INSERM U1227 Lymphocytes B et Autoimmunite (LBAI)},%Department and Organization
            state={Brest},
            country={France}}

\author[inst1,inst2]{Rachid Zeghlache}
\author[inst1,inst3]{Pierre-Henri Conze}
\author[inst1,inst2]{Mostafa El Habib Daho}
\author[inst1,inst2]{Yihao Li}
\author[inst1,inst2]{Alireza Rezaei}
\author[inst5]{Hugo Le Boité}
\author[inst5]{Ramin Tadayoni}
\author[inst5]{Pascal Massin}
\author[inst1,inst2,inst4]{Béatrice Cochener}
\author[inst1,inst2,inst6]{Ikram Brahim}
\author[inst1]{Gwenolé Quellec}
\author[inst1,inst2]{ Mathieu Lamard}

% \def\keyFont{\fontsize{8}{11}\helveticabold }
% \def\firstAuthorLast{Zeghlache {et~al.}} %use et al only if is more than 1 author
% \def\Authors{Rachid Zeghlache\,$^{1,2,*}$, Pierre-Henri Conze$^{1,3}$, Mostafa El Habib Daho$^{1,2}$, Yihao Li$^{1,2}$, Hugo Le Boité $^{5}$, Ramin Tadayoni $^{5}$, Pascal Massin $^{5}$, 
%  Béatrice Cochener $^{1,2,4}$, 
% Ikram Brahim $^{1,2,6}$, 
% Gwenolé Quellec $^{1}$, 
% and Mathieu Lamard $^{1,2}$}
% % Affiliations should be keyed to the author's name with superscript numbers and be listed as follows: Laboratory, Institute, Department, Organization, City, State abbreviation (USA, Canada, Australia), and Country (without detailed address information such as city zip codes or street names).
% % If one of the authors has a change of address, list the new address below the correspondence details using a superscript symbol and use the same symbol to indicate the author in the author list.
% \def\Address{$^{1}$ LaTIM UMR 1101, Inserm, Brest, France  \\
% $^{2}$ University of Western Brittany, Brest, France \\
% $^{3}$ IMT Atlantique, Brest, France \\
% $^{4}$ Ophtalmology Department, University Hospital of Brest, Brest, France\\
% $^{5}$ Lariboisière Hospital, AP-HP, Paris, France \\
% $^{6}$ INSERM U1227 Lymphocytes B et Autoimmunite (LBAI) Brest, France

\begin{abstract}
%% Text of abstract
Pre-training strategies based on self-supervised learning (SSL) have proven to be effective pretext tasks for many downstream tasks in computer vision. Due to the significant disparity between medical and natural images, the application of typical SSL is not straightforward in medical imaging. Additionally, those pretext tasks often lack context, which is critical for computer-aided clinical decision support. In this paper, we developed a longitudinal masked auto-encoder (MAE) based on the well-known Transformer-based MAE. In particular, we explored the importance of time-aware position embedding as well as disease progression-aware masking. Taking into account the time between examinations instead of just scheduling them offers the benefit of capturing temporal changes and trends. The masking strategy, for its part, evolves during follow-up to better capture pathological changes, ensuring a more accurate assessment of disease progression. Using OPHDIAT, a large follow-up screening dataset targeting diabetic retinopathy (DR), we evaluated the pre-trained weights on a longitudinal task, which is to predict the severity label of the next visit within 3 years based on the past time series examinations. Our results demonstrated the relevancy of both time-aware position embedding and masking strategies based on disease progression knowledge. Compared to popular baseline models and standard longitudinal Transformers, these simple yet effective extensions significantly enhance the predictive ability of deep classification models.
\end{abstract}

\begin{keyword}
%% keywords here, in the form: keyword \sep keyword
Diabetic retinopathy \sep  self-supervised learning
%% PACS codes here, in the form: \PACS code \sep code
longitudinal analysis \sep pretext task
%% MSC codes here, in the form: \MSC code \sep code
%% or \MSC[2008] code \sep code (2000 is the default)
 \sep disease progression
\end{keyword}

\end{frontmatter}

%% \linenumbers

%% main text

\section{Introduction}

\begin{figure}[t]
\includegraphics[width=\textwidth]{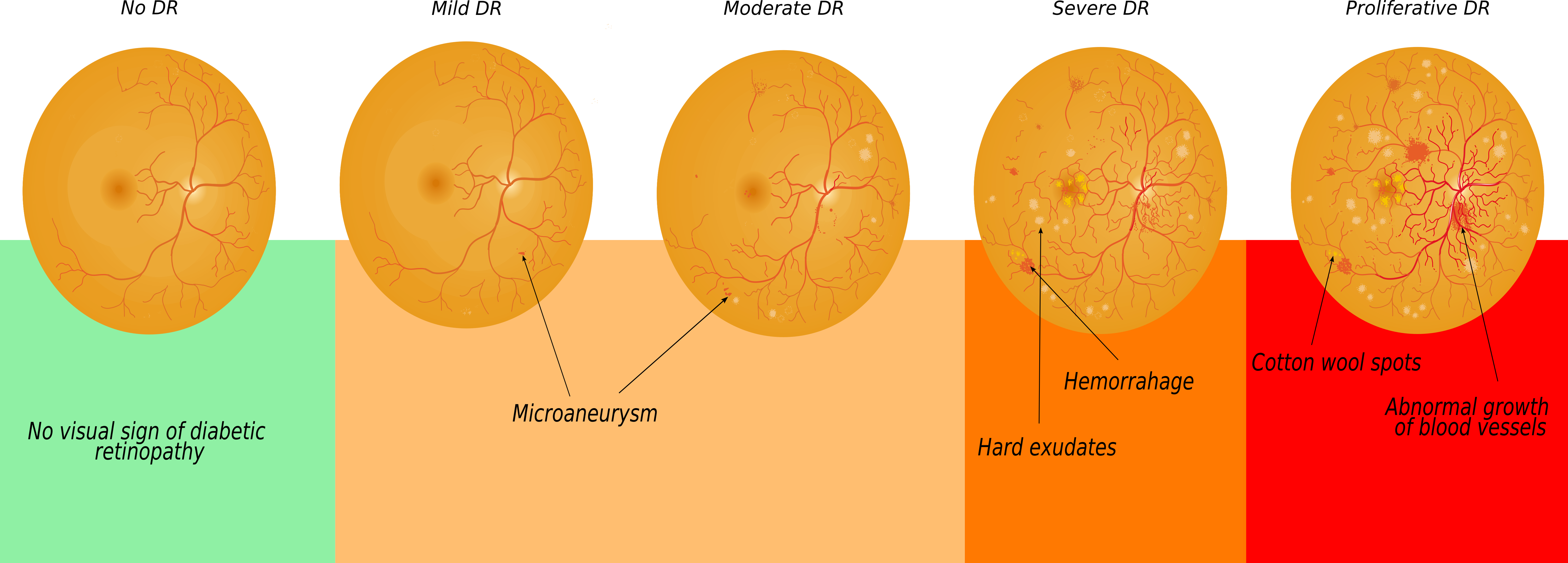} 
\caption{Diabetic retinopathy progression stages. The number and type of anomalies increase with severity. As severity increases, we observe more anomalies in peripheral zones.} 
\label{fig:diabetic_retinopathy_progression}
\end{figure}

Diabetic retinopathy (DR), a leading cause of vision loss worldwide, affects millions of people with diabetes \cite{kroppDiabeticRetinopathyLeading2023}. DR is a common condition in individuals with diabetes and is a significant threat to vision worldwide. According to the International Diabetes Federation, it is estimated that DR will affect more than 700 million people by 2045 \cite{Saeedi2019}. DR occurs when retinal blood vessels are damaged by long-term high blood sugar levels, leading to leakage and swelling. Detecting DR early through retinal imaging, especially by identifying red and bright lesions, is crucial for preventing vision loss. Color fundus photographs (CFP) are commonly used to assess the structure of the retina and identify lesions such as microaneurysms, hemorrhages, and exudates. The international clinical DR severity scale divides DR into five grades, ranging from no DR (grade 0) to severe proliferative DR (grade 4). Timely interventions can effectively manage the early-to-middle stages of DR (grades 1, 2, 3), which are characterized by progressive damage to the small blood vessels and blockages. Proliferative DR (grade 4) carries a high risk of vision loss due to significant hemorrhages (Fig.~\ref{fig:diabetic_retinopathy_progression}). Detecting DR early and providing appropriate treatment, especially during the mild to moderate stages of non-proliferative DR (NPDR), can significantly delay the progression of DR and prevent vision problems related to diabetes. The existing guidelines for DR screening suggest that everyone with diabetes should undergo annual exams, regardless of their specific risk factors. However, this procedure could result in missed diagnoses because it does not take into account the variation in individual risk. Utilizing artificial intelligence (AI) shows potential for enhancing DR screening as it allows the creation of models that can predict an individual's likelihood of developing or advancing DR \cite{rajeshArtificialIntelligenceDiabetic2023,almattarDiabeticRetinopathyGrading2023}. 

Despite the potential for AI to bring a revolutionary change in DR screening, there are still several important areas of research that need to be addressed. One such area is the limited scope of existing AI models, which mainly concentrate on predicting the risk of DR within a two-year timeframe based solely on a single color fundus photograph \cite{Bora2021,Rom2022,nderituPredictingProgressionReferable2022}. However, DR is a chronic disease that can progress gradually over time, emphasizing the need for models that can predict risk over extended periods, as proposed in \cite{daiDeepLearningSystem2024}. This paper presents a methodology where a pre-training step on a large dataset is combined with a deep survival analysis model. The goal is to predict the progression of DR over a period of 1 to 5 years using only a single CFP.

Additionally, previous methods have focused solely on single-color fundus photography and ignored previous examination information in their deep learning process. The reliance on a single CFP and the disregard for previous examination data in the prediction of DR raise several concerns. Moreover, CFP alone may not capture the subtle changes indicative of DR progression, especially in the early stages. Additionally, neglecting previous examination information overlooks the longitudinal dynamics of DR, potentially leading to inaccurate risk assessment and inadequate screening intervals. Incorporating both CFP and longitudinal data into risk prediction models could enhance their accuracy and effectiveness for personalized DR screening.

Recent research suggests that self-supervised learning can learn efficient representations by training on pretext tasks before resolving downstream supervised tasks (e.g., classification tasks) and may tackle the challenge of limited label availability. Current models are typically based on contrastive learning, which consists of learning representations by making the models distinguish the differences and similarities between samples. Previous studies have generated or identified similar or dissimilar sample pairs (also known as positive and negative pairs) based on data augmentation \cite{SIMCLR} or sample organization in a lookup dictionary \cite{Moco}. The use of standard SSL in medical imaging is not straightforward due to the notable disparities between medical and natural images. Medical images differ from natural images in terms of content, data format, noise and artifacts, dimensionality, semantic meaning, and applications. They are employed to identify diseases, observe the response to treatment, plan and guide surgical operations. Also, medical images are usually disturbed by noise and artifacts, while natural images are generally less noisy. Finally, medical images have a more precise semantic significance compared to natural images, whose semantic content may be more ambiguous.

Furthermore, context, which is necessary for computer-aided clinical decision assistance, is frequently missing from such pretext tasks. In this direction, some papers have used SSL in the context of DR to learn better representation to detect DR using more context in the design of their pretext task. Authors in \cite{huang2021lesionbased} have used the SimCLR framework \cite{SIMCLR} directly on the sub-region of color fundus photographs which contain pathological tissues. Authors in \cite{zeghlache} have compared the use of longitudinal pretext tasks to study longitudinal changes between consecutive pairs. Using more context in the pretext task's definition enables methodologies developed in \cite{zeghlache,huang2021lesionbased} to get enriched representation to solve their target classification task.

Transformers \cite{vaswani2017attention} are overpowering the old convolutional neural network (CNN) paradigm in medical imaging \cite{shamshad2022transformers}. In particular, masked auto-encoder (MAE), a Transformer-based method \cite{devlin2019bert}, is another type of promising self-supervised learning approach. Introduced in the context of natural language processing (NLP) with BERT \cite{devlin2019bert}, its objective is to reconstruct a masked version of its input using an auto-encoder. MAEs have become well-known for their capacity to draw out strong and transferable representations from unlabeled data, their self-supervised learning approach, and their success in capturing global context and semantic relationships. These characteristics make MAEs advantageous for pre-training deep learning models and improving their execution in machine learning applications. Later, the same idea was extended to images in \cite{he2021masked} with vision Transformers (ViT) encoder \cite{dosovitskiy2021image}. For action recognition from videos \cite{HE201934,
LI2023104740,
LI2021104244,
ZHANG2023104744}, different extensions of the MAE framework applied to spatial-temporal data have been proposed  \cite{feichtenhofer2022masked}.

In this particular situation, the author intends to fully leverage an extensive dataset of retinal images through the application of Self-Supervised Learning (SSL) as outlined in Zhou et al.'s work on foundation models \cite{zhouFoundationModelGeneralizable2023a}. Specifically, the proposed model, RETFound, which is based on the Masked Autoencoder (MAE) approach,  demonstrates superior performance in disease detection and prognosis tasks compared to conventional models. The unique approach of RETFound involves utilizing self-supervised learning on a vast dataset of unlabeled retinal images to extract representations that can be applied in various contexts. Foundation models are specifically designed to acquire general-purpose representations of data. These models are trained on extensive amounts of unlabeled data and can be employed to address a wide range of subsequent tasks. When faced with limited labeled data, foundation models can still be effective as they learn features that are relevant to diverse tasks. On the other hand, specialized pretext tasks are tailored to learn task-specific representations. These tasks are typically trained on smaller amounts of labeled data and can be more effective when dealing with a large amount of labeled data. In this paper, we choose to utilize a specialized pretext task to extract features that are specifically relevant to the progression of DR. This decision is motivated by the fact that specialized pretext tasks can be more efficient to train compared to foundation models.

Since their introduction, one strategy to further improve the capacity of MAE involves the forethought of specific masking strategies, either based on handcrafted methods or learned during training \cite{chen2023improving,mao2023medical}. In the context of early-stage DR, the ophthalmologist pays closer attention to the center of the CFP to detect lesions, namely microaneurysms. And the more the disease progresses, the more the periphery is taken into consideration for DR severity assessment \cite{WILKINSON20031677}. For early stages, the center of the image is critical for DR severity assignment, while for later stages, the center of the image is important but less discriminant than the presence of lesions in the periphery \cite{STINO2023}. Moreover, lesions induced by DR present different sizes according to the stage of the disease \cite{Alyoubi2021DiabeticRF} (Fig.~\ref{fig:diabetic_retinopathy_progression}).

From this observation, we developed a masking strategy that masks areas according to the severity progression. Thus, we aim at forcing the model to focus on likely discriminatory areas according to the DR severity. In addition, positional encoding is used in Transformers to inject order into the sequences. However, it can only represent simple sequential orders, and there is no notion of time between consecutive examinations, which is crucial. For instance, the examination frequency indicates the disease's underlying progression. With the purpose of taking advantage of such information, time-aware models are a key methodology. They were first introduced with the time-LSTM (T-LSTM) \cite{Baytas}. Using time-aware models to tackle disease progression using irregular time series images has proven effective. Zeghlache et al. \cite{zeghlache2023} have used classical versus T-LSTM to predict the severity of the last examination in the context of DR using time series images. Their results demonstrated the usefulness of time-aware LSTM compared to standard LSTM.  We propose a straightforward modification of the ViVit transformer and introduce a Time-ViVit by adding a time-aware position encoding along the temporal dimension after the embeddings layer in order to make model time dependant.

In summary, our contributions are as follows:

\begin{enumerate}
    \item We introduce a longitudinal masked auto-encoder (L-MAE) to overcome the aforementioned limitations while still getting benefits from the MAE framework. More specifically, we add a time-aware embedding to make the embedding aware not only of the sequence order but also of its duration and relative position in time.

    \item  We also integrate a new masking strategy to exploit and inject domain knowledge with the hope of producing a better representation for longitudinal downstream tasks. 

    \item Superior performance on a longitudinal DR progression task when using our pre-trained weights compared to baseline methods.

    \item To the best of our knowledge, this work is the first to use a L-MAE with an emphasis on label-aware masking and time-aware embedding for disease progression purposes.
\end{enumerate}

\section{Materials and methods}

\subsection{Temporal position embedding}

Sequential models, like Recurrent Neural Networks (RNN) or Transformers, do not consider the time span between events and thus capture sequential signals rather than temporal patterns. By construction, recurrent networks structure their inputs by sequences. However, self-attention recognizes event orderings via positional encoding. The positional encoding (PE) was introduced in \cite{vaswani2017attention} for Transformers. The PE is a vector of $d$ dimensions that provide a relative position about a certain element in the sequence. Let $k$ be the position in the input sequence that is desired, $p_k$ its related encoding in $\mathbb{R}^d$, and $d$ the encoding dimension. Let $f :\mathbb{N} \rightarrow \mathbb{R}^d$ be the mapping that constructs the output vector $p_k$, as follows:

\begin{align}
  p_{k}^{(i)} = f(k)^{(i)} & = 
  \begin{cases}
      \sin({\omega_l} . k),  & \text{if}\  i = 2l \\
      \cos({\omega_l} . k),  & \text{if}\  i = 2l + 1
  \end{cases}
\label{eq:p_embedding}
\end{align}

% \noindent The vector is calculated such that it is a geometric progression from $2\pi$ to $10000 \times 2\pi$ in terms of wavelengths. 

This vector is then added to all tokens in the sequence, depending on their relative position $k$ and the dimension of the vector $i$. The value of $\omega_l$ is $\frac{1}{10000^{2l/d}}$.

\begin{figure}[H]
\includegraphics[width=\textwidth]{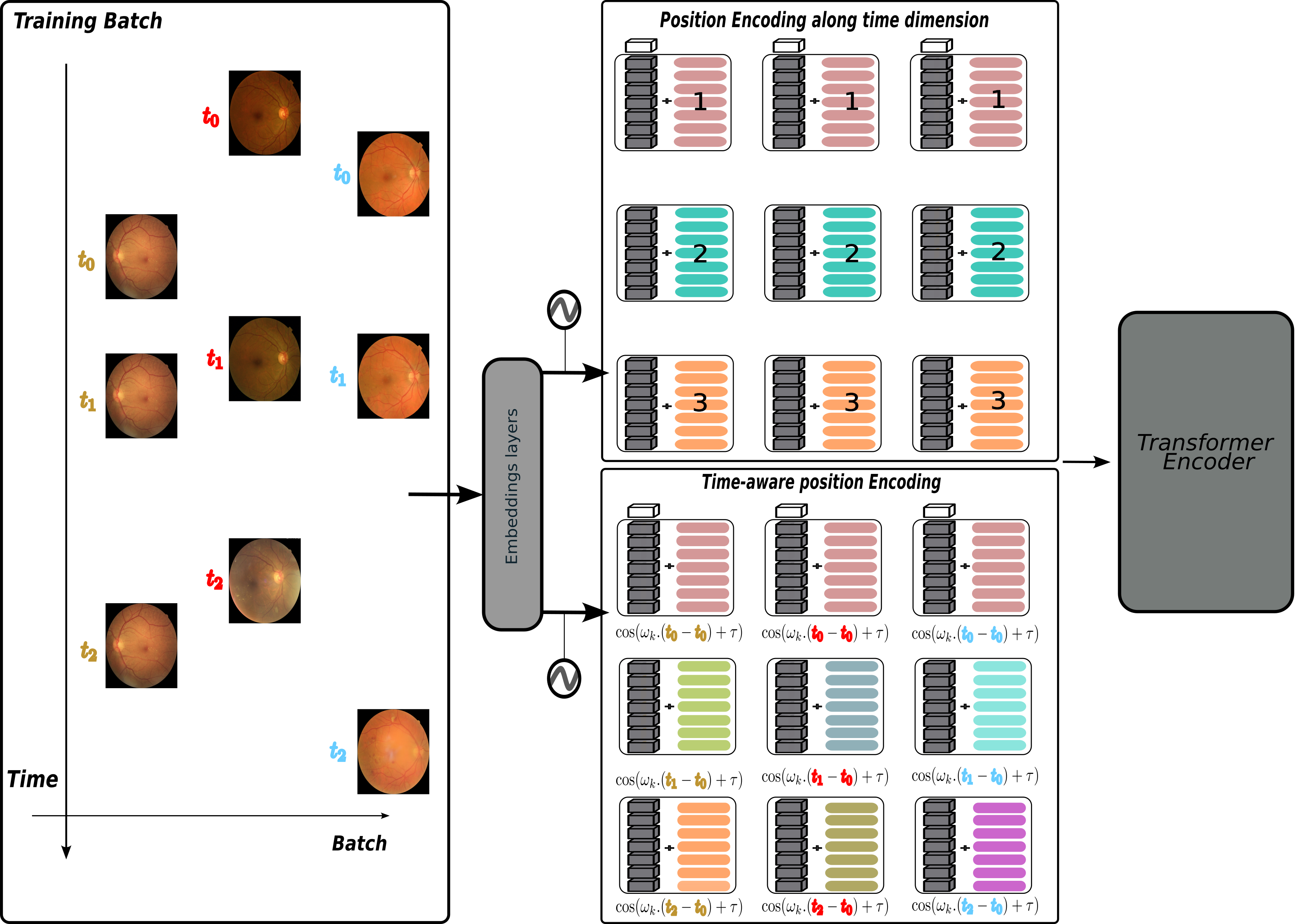} 
\caption{An example of position encoding in the time dimension is provided. The conventional approach to position encoding only encodes the order of positions, disregarding the time intervals between elements. Our innovative masking strategies, which are sensitive to disease progression dynamics, incorporate time-dependent information to allow the model to capture the dynamics of disease progression.} 
\label{fig:time_aware_encoding_}
\end{figure}

In order to induce temporal knowledge into sequence tokens and inspired by the authors of \cite{Lin2021PretrainingCA, liTimedistanceVisionTransformers2023}, we present a functional feature map that embeds time spans into high-dimensional spaces. Named \textit{time-aware PE}, it is defined with the following expression: 
\begin{align}
p_{time-aware}^{(i)} = f(t_{i})^{(i)} & := \cos({\omega_k}.( t_{i}-t_{0} )+ \tau)
\label{eq:t_p_embedding}
\end{align}

\noindent with $\omega_k$ and $\tau$ are the learnable parameters of the time-aware embedding, with $t_{0}$ as the starting date of the sequence of examination and $t_{i}$ represent the absolute time (in years in our case) for sequence tokens. A sequence is composed of multiple tokens. For a given sequence, we add this temporal encoding embedding to all tokens that share the same absolute time for a given sequence. Fig.~\ref{fig:time_aware_encoding_} highlights the difference between the two position encodings.

\subsection{Vision Transformer}

The traditional Transformer model works with a sequence of token embeddings in one dimension. In order to handle two-dimensional images, the input image $\mathbf{x} \in \mathbb{R}^{H \times W \times C}$ is reshaped into a sequence of flattened two-dimensional patches $\mathbf{x}_p \in \mathbb{R}^{N \times (P^2 \cdot C)}$. Here, $(H, W)$ represents the spatial dimensions of the original image, $C$ represents the number of channels, $(P,P)$ represents the spatial resolution of each image patch, and $N=HW/P^2$ represents the resulting number of patches. This number of patches also serves as the effective input sequence length for the Transformer. The Transformer uses a fixed latent vector size $D$ for all its layers. Therefore, the patches are flattened and transformed into $D$ dimensions using a trainable linear projection. (Eq.~\ref{eq:embedding}). 

As performed in BERT \cite{devlin2019bert}, a trainable embedding is added to the sequence of embedded patches ($\mathbf{z}_0^0=\mathbf{x}_\text{class}$), whose state at the output of the Transformer encoder ($\mathbf{z}^0_L$) serves as the image representation $\mathbf{y}$ (Eq.~\ref{eq:final_rep}).

A classification head is attached to $\mathbf{z}^0_L$ to perform the prediction. This head is implemented by a multi-layer perceptron (MLP) with one hidden layer. Position embeddings (Eq.~\ref{eq:p_embedding}) are added to the patch embeddings to retain positional information. The Transformer encoder \cite{vaswani2017attention} consists of alternating layers of Multiheaded Self-Attention (MSA) and MLP blocks (Eq.~\ref{eq:msa_apply}, \ref{eq:mlp_apply}). Layernorm (LN) is applied before every block, and residual connections after every block. The MLP contains two layers with a Gaussian Error Linear Unit (GELU) non-linearity. A schematic representation of the Transformer architecture can be seen in Fig.\ref{fig:proposed_method}.
\begin{align}
    \mathbf{z}_0 &= [ \mathbf{x}_\text{class}; \, \mathbf{x}^1_p \mathbf{E}; \, \mathbf{x}^2_p \mathbf{E}; \cdots; \, \mathbf{x}^{N}_p \mathbf{E} ] + \mathbf{E}_{pos},
    && \mathbf{E} \in \mathbb{R}^{(P^2 \cdot C) \times D},\, \mathbf{E}_{pos}  \in \mathbb{R}^{(N + 1) \times D} \label{eq:embedding} \\
    \mathbf{z^\prime}_\ell &= {MSA}({LN}(\mathbf{z}_{\ell-1})) + \mathbf{z}_{\ell-1}, && \ell=1\ldots L \label{eq:msa_apply} \\
    \mathbf{z}_\ell &= {MLP}({LN}(\mathbf{z^\prime}_{\ell})) + \mathbf{z^\prime}_{\ell}, && \ell=1\ldots L  \label{eq:mlp_apply} \\
    \mathbf{y} &= {LN}(\mathbf{z}_L^0)
    \label{eq:final_rep}
\end{align}

\subsection{Progression-aware masking strategies}

\begin{figure}[H]
\includegraphics[width=\textwidth]{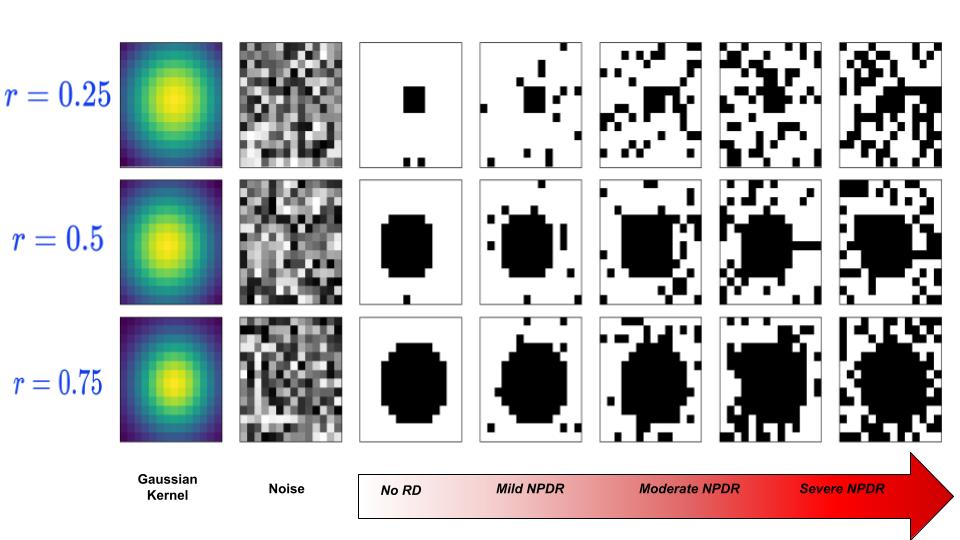} 
\caption{Illustration of our proposed progression-aware masking strategies.} 
\label{fig:overview}
\end{figure}

Our proposed masking strategies are based on the computation of a spatial Gaussian kernel $(a_{ij}$) and random uniform noise ($b_{ij}$). The circular form is produced by the spatial Gaussian kernel, while the random noise is present to propagate the mask aligned with the increase of severity (Fig.~\ref{fig:overview}). Firstly, based on the size of the image, we note $L=H=W$, and we create a grid based on the $q=\sqrt{L}$. We then define a point $c = (c_x,c_y)$ as the center of this grid size that will be the center of the 2D Gaussian. When creating to increase the diversity of the masking strategy, an offset of $\pm$ 1 in each direction is randomly applied to $c$.
\begin{equation}
a_{ij} = exp\left(\frac{-\pi}{r q^2}{\frac{(i-c_x)-(j-c_y)}{2}}^2\right)
\label{eq:labelawaremasl}
\end{equation}   
\begin{equation}
b_{ij} \stackrel{iid}{\sim} U[0,1], \: \text{where} \: i=j \in [1:q]  \label{eq:uniform_matrix}
\end{equation}

To create our mask, we used $a_{ij}$  (Eq.~\ref{eq:labelawaremasl}) and $b_{ij}$  (Eq.~\ref{eq:uniform_matrix}) and binarized them using the mentioned thresholds. With $r \in [0,1]$ defining the intensity of the Gaussian kernel. The Gaussian spatial kernel smoothly goes down when the pixel distance becomes larger with the defined center $c$. In our experiments, the value of $r$ is 0.75. To mask more regions and propagate the mask according to the severity progression, we define the following thresholds: $T_{severity} = 1-0.1 \times s_{DR}$ with $s_{DR}\in$ [0,1,2,3,4] referring to the DR grades. The first mask is defined with $ m_{a_{ij}} =(a_{ij} < r)$ and the second mask is defined by  $ m_{b_{ij}} = (b_{ij} < T_{severity})$. Finally, we obtain our proposed mask by multiplying the two masks. Fig.~\ref{fig:overview} illustrates the masking scheme based of the value of $r$ and $T_{severity}$.

\begin{equation}
m_{ij} = m_{a_{ij}} * m_{b_{ij}}
\label{eq:labelawaremasl_final}
\end{equation}

\subsection{Longitudinal masked auto-encoder (L-MAE)}

\begin{figure}[H]
\includegraphics[width=\textwidth]{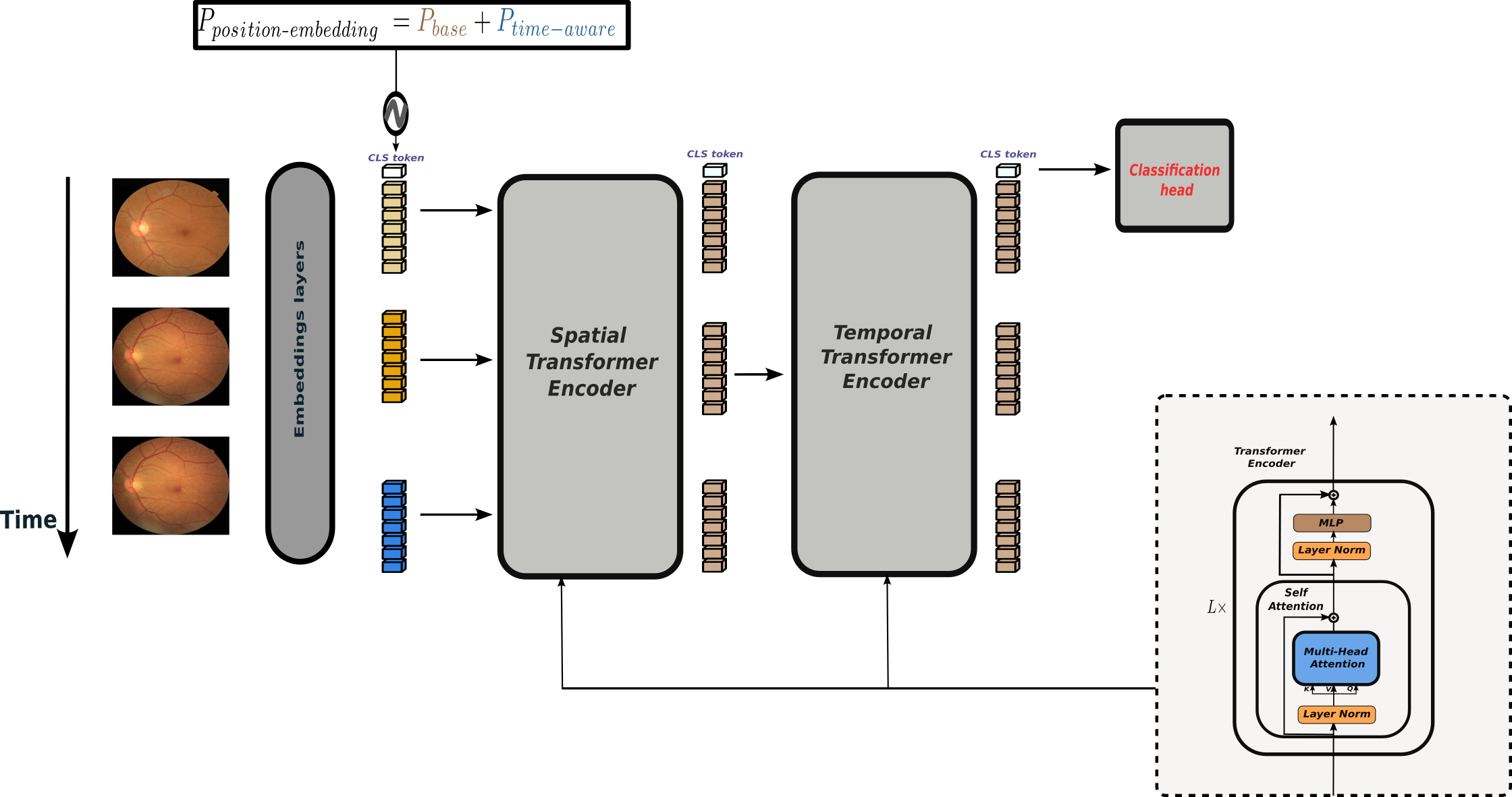}

\caption{Illustration of the ViVit. In the representation, we include our time-aware position encoding instead of a regular position encoding along the temporal dimension.}
\label{fig:ViVit}
\end{figure}

Our encoder is a basic Video Vision Transformer (ViViT) \cite{arnab2021vivit} that only operates on the visible set of embedded patches ( i.e., not on the masked patches), inspired by standard ViT \cite{dosovitskiy2021image}. An illustration of the ViVit is presented Fig.~\ref{fig:ViVit}. This design significantly reduces computational and memory demands, resulting in a more practical solution.

Our decoder is another standard ViT that processes the combined set of encoded patches and mask tokens. It also incorporates decoder-specific positional embeddings and temporal embeddings. The decoder is constructed to be more compact than the encoder. Even though it processes the entire set, its complexity is less than the encoder's. As performed in \cite{feichtenhofer2022masked}, the decoder predicts only the patches in the pixel space. The encoder and decoder are independent of the problem's space-time structure. There is no hierarchy or spacetime factorization. Rather than the usual approaches, this technique utilizes global self-attention to acquire valuable information from data. Our L-MAE takes a sequence of Time-Series Images (TSI), $S \in T \times R^3 \times H \times W$, as input. A 3D convolution layer is used to construct the patch embedding. Following the implementation of \cite{he2021masked}, our Masked Autoencoder (MAE) is a straightforward AE approach that reconstructs the supplied TSI based on the masked TSI. The decoder of the MAE is fed with the full set of tokens, which includes encoded visible patches and mask tokens. (Fig.~\ref{fig:overview}). Each mask token is a shared, learned vector that indicates the presence of a patch that needs to be predicted. We incorporate positional embeddings (for the spatial order) and temporal embeddings into all tokens in this complete set. The loss function is analogous to the one used in a regular MAE:

\begin{figure}[H]
\includegraphics[width=\textwidth]{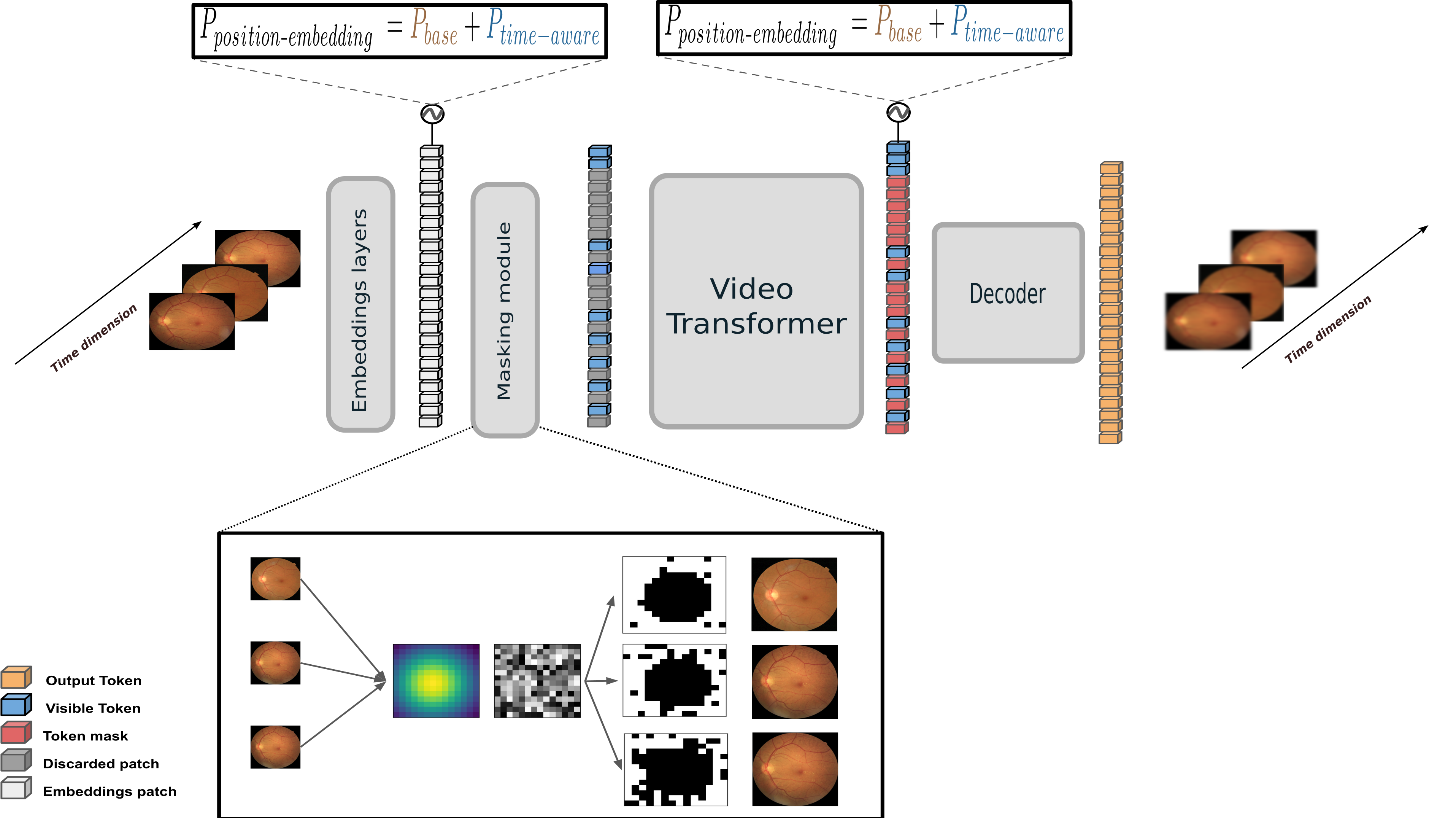}

\caption{Illustration of our longitudinal masked auto-encoder. Our proposed longitudinal masked autoencoder differs from the classic Video-MAE from two perspectives. The first one is the fact that, in the embedding layers, we add our proposed time-aware position embedding to the classical position embeddings used in the transformer. The second difference lies in the masking strategies.}
\label{fig:proposed_method}
\end{figure}

\begin{equation}
L = \sum_{p\in\chi} \parallel S(p) - \tilde{S}(p)\parallel_2^2 \end{equation}
 
\noindent where $p$ is the token index, $\chi$ the set of masked tokens (random, full frame masked or progression-aware as in Fig.\ref{fig:overview}), $S$ the image sequence and $\tilde{S}$ the reconstructed image sequence.

\subsection{Dataset}

The proposed models were trained and evaluated on OPHDIAT \cite{ophdiat}, a large CFP database collected from the Ophthalmology Diabetes Telemedicine network consisting of examinations acquired from 101,383 patients between 2004 and 2017.Out of over 763,848 interpreted CFP images, around 673,000 were assigned a diabetic retinopathy (DR) severity grade, while the rest were deemed ungradable. The patients' ages ranged from 9 to 91. Each examination included at least two images per eye, and most patients had multiple images with different fields of view for both eyes. For easier selection, one image per eye was randomly chosen for each examination. The dataset was labeled according to the international clinical DR severity scale (ICDR)\cite{WILKINSON20031677}. From the OPHDIAT database, we selected patients who had at least four visits to perform our experiments. This resulted in 7244 patients and 13997 sequences. This dataset was further divided into training (60$\%$), validation (20$\%$), and test (20$\%$) based on patients. 

\subsection{Target task}

The prediction task was to predict the severity grade of the next exam in three years.  Conventional screening guidelines, advocating annual retinal examinations for all individuals with diabetes \cite{dasRecentlyUpdatedGlobal2021}, fail to consider individual risk profiles, resulting in unnecessary examinations and potential missed diagnoses. This approach is inefficient as most patients with DR do not exhibit significant disease progression within two years, highlighting the protracted nature of DR development \cite{kroppDiabeticRetinopathyLeading2023}. Moreover, the three-year span was also chosen because it was the interval that contained the largest number of patients with a known development of diabetic retinopathy within a period of time.
\subsection{Implementation details}

Our Time-ViViT encoder is based on the implementation introduced in \cite{arnab2021vivit}. Our encoder and decoder follow the vanilla ViViT architectures presented in the paper \cite{arnab2021vivit}. Specifically, the architecture extended to work with multiple frames based on the ViT-B \cite{dosovitskiy2021image} for simplicity. We used three different image sizes, $224 \times 224$, $256 \times 256$, $448 \times 448$, and tried two different sizes of the patch: $ 16 \times 16 $ and $32 \times 32$. We used a temporal patch size of 1 since the images are not temporally aligned. Three different masking ratios (denoted $r_{mask} \in$ \{25\%,50\%,75\%\}) were used when the random masking strategy was performed. The different networks were trained for 400 epochs by the AdamW optimizer with a learning rate of $5 \times 10^{-3}$, OneCycleLR as a scheduler, and a weight decay of $10^{-5}$, using an NVIDIA A6000 GPU with the PyTorch framework. For pre-training the L-MAE, we used three different types of PE: an empty encoding equivalent to no position embedding added for the temporal dimension, base encoding equivalent to the PE in Eq.~\ref{eq:p_embedding} and time-aware embedding based on Eq.~\ref{eq:t_p_embedding}.

\subsection{Evaluation of the feature extractor}

To assess the effectiveness of the features extracted by the different pre-training strategies, we only used \textbf{\textit{fine-tuning}} on the longitudinal downstream task, which was to predict the severity of the next visit within a 3-year period based on previous examinations. The model was trained using cross-entropy loss and the Area Under the receiver operating characteristic Curve (AUC) was used to evaluate the models with three binary tasks: predicting at least non-proliferative diabetic retinopathy (NPDR) in the next visit (\textbf{AUC Mild+}), at least moderate NPDR in the next visit (\textbf{AUC Moderate+}) and finally at least severe NPDR in the next visit (\textbf{AUC Severe+}). The fine-tuning scenario consisted of initializing the Time-ViViT (encoder) weights using the weights obtained by training the L-MAE and the time-aware embeddings. These embedding layers were learned during the pretraining. After the encoder, we added an MLP layer to predict the label of the next visit. In the \textbf{\textit{fine tuning}} scenario, the model was trained during 200 epochs, with a learning rate of 0.001 and weight decay of 0.0001.

\subsection{Methods comparaison}

We compared our approach to classic and state-of-the-art techniques that process sequence images and the potential use of time. We trained each model using the cross-entropy loss and monitored the model's performance on the validation set by tracking the loss value. The best model was then retained.

\noindent \textbf{2D CNN+pooling} \cite{CNN+late+pooling} where each frame was fed to the CNN, thus producing latent vectors. These vectors were concatenated, and a global average pooling was applied on top of the formed representation, then injected in a classification head to predict the label of the next visit.

\noindent \textbf{CNN+LSTM} \cite{wang2016cnnrnn} where each frame was fed to the CNN and where each frame's latent vector was then sequentially fed to an LSTM layer. The last representation produced by the LSTM was given to the classification head to predict the label of the next visit.

\noindent \textbf{CNN+T-LSTM} was employed similar to the last method, but instead of the classic LSTM, we used time-LSTM introduced by \cite{Baytas}. We used the Pytorch implementation of \cite{zeghlache2023} in order to have a model that is coupled with CNN and the use of time. 

\noindent \textbf{Video ViT (ViViT)}   \cite{arnab2021vivit} was used to classify image sequences. In their paper, they proposed three variants. In our paper, we implement the first version of ViVit, which extends ViT to include the capability to model the relationships between all spatiotemporal tokens in pairs.

\noindent \textbf{Time-ViViT}. We extend the previous model with our time-aware embeddings, motivated by the fact that previous study \cite{Lin2021PretrainingCA,thomas} have demonstrated the effectiveness of including time-related embeddings instead of just token position along the temporal axis.

\section{Results}

According to our experiments, we observed that the image size was not important for the Transformer. But instead, what was more discriminant was the number of total tokens (i.e., the number of patches). The opposite was observed for the CNN-based method. The first three rows in Tab.~\ref{table:results_comparaison} report the baselines. The next three lines present the Time-ViViT trained from scratch but with different temporal embedding. We also observe that the classical PE is an important component for the Transformer-based approach because the model with empty encoding has poor predictive ability compared to the baseline or Time-ViViT with other embeddings, the same being observed across the different masking strategies.

In addition, using the temporal positional embedding compared to empty or base embedding or baselines exhibits the best performance for the \textbf{AUC Severe+}. This suggests that our temporal PE provides key information to the model. Similarly, we observed that the CNN+T-LSTM demonstrates a better ability to predict disease progression than the standard CNN+LSTM. This is aligned with the fact that adding temporal information provides better context for predicting the disease progression \cite{Bridgee000569,zeghlache2023}. Except for the CNN baselines, the best performance for \textbf{AUC Mild+} and \textbf{AUC Moderate+} is obtained using our masking strategy and the time-aware embedding. 

The rest of the results presented in Tab.~\ref{table:results_comparaison_hp_masking} concern using different masking strategies in conjunction with a specific position embedding in the temporal dimension. Concerning the masking strategy, we report that masking more areas enables the MAE to capture a better representation, yielding a better performance across the different binary tasks.

\begin{table}[H]
\centering
\caption{Comparison of approaches in terms of the different AUC scores for the predict next task, on the test set. We only report the best performance of the presented methods for the different hyper-parameters. Results in bold show the best results.}
\label{table:results_comparaison}

\resizebox{\columnwidth}{!}{%

\begin{tabular}{|c|c|c|c|c|c|c|c|c|} 
\hline
Method                                                      & \begin{tabular}[c]{@{}l@{}}Masking\\ parameter\end{tabular} & \begin{tabular}[c]{@{}l@{}}Temporal\\ embedding\end{tabular} & \begin{tabular}[c]{@{}l@{}}Masking\\ strategies\end{tabular} & \begin{tabular}[c]{@{}l@{}}Pretrain\\ weights\end{tabular} & Backbone & \begin{tabular}[c]{@{}l@{}}AUC\\ Mild+\end{tabular} & \begin{tabular}[c]{@{}l@{}}AUC\\ Moder+\end{tabular} & \begin{tabular}[c]{@{}l@{}}AUC\\ Severe+\end{tabular}  \\ 
\hline

\begin{tabular}[c]{@{}l@{}}2D CNN+\\ pooling\\\end{tabular} & -                                                           & -                                                            & -                                                            & ImageNet                                                   & ResNet50 & 0.543                                               & 0.597                                                & 0.668                                                  \\ 
\hline
\begin{tabular}[c]{@{}l@{}}CNN+\\ LSTM\\\end{tabular}       & -                                                           & -                                                            & -                                                            & ImageNet                                                   & ResNet50 & 0.566                                               & \textbf{ 0.618}                                      & 0.739                                                  \\ 
\hline
\begin{tabular}[c]{@{}l@{}}CNN+\\ T-LSTM\\\end{tabular}     & -                                                           & -                                                            & -                                                            & ImageNet                                                   & ResNet50 & \textbf{0.595}                                      & 0.614                                                & 0.783                                        \\  
\hline
 \begin{tabular}[c]{@{}l@{}}ViViT \end{tabular}& -& -& -& -& ViT& 0.496	& 0.541 & 0.556 \\ 
 \hline 
 \begin{tabular}[c]{@{}l@{}}ViViT \end{tabular}& $r = 0.75 $& -& Prog-aware& L-MAE& ViT & 0.566 & 0.590 & 0.744 \\ 
\hline
\begin{tabular}[c]{@{}l@{}}L-ViViT \\\end{tabular}                                                    & $r = 0.75 $                                                 & Time-aware                                                   & Prog.-aware                                                  & L-MAE                                                      & ViT      & 0.560                                      & 0.584                                       & \textbf{0.797}                                         \\
\hline
\end{tabular}%
}
\end{table}

\subsection{Ablation study}

In this section, we further study the impact of the specific hyperparameters and components presented in our proposed approach. We tested the efficacy of the proposed method by comparing it to the baseline method and different variants.

\begin{table}
\centering
\caption{Comparison of approaches in terms of the different AUC scores for the predict next task, on the
test set regarding the importance of masking hyper-parameters. }
\label{table:results_comparaison_hp_masking}

\resizebox{\columnwidth}{!}{%

\begin{tabular}{|l|c|c|l|l|l|l|} \hline  

Method & Masking parameter& \begin{tabular}[c]{@{}c@{}}Temporal\\ embedding\end{tabular} & \multicolumn{1}{|c|}{\begin{tabular}[c]{@{}c@{}}Masking\\ strategies\end{tabular}} & \multicolumn{1}{|c|}{\begin{tabular}[c]{@{}c@{}}AUC\\ Mild+\end{tabular}} & \multicolumn{1}{|c|}{\begin{tabular}[c]{@{}c@{}}AUC\\ Moder+\end{tabular}} & \begin{tabular}[c]{@{}l@{}}AUC\\ Severe+\end{tabular}  \\ \hline  

ViViT& $r_{mask}=$ 0.25                                              & Empty                                                        & Random                                                                            & 0.500                                                                    & 0.528                                                                     & 0.454                                                  \\ \hline  

ViViT& $r_{mask}=$ 0.5                                               & Empty                                                        & Random                                                                            & 0.482                                                                    & 0.530                                                                     & 0.481                                                  \\ \hline  

ViViT& $r_{mask}=$ 0.75                                              & Empty                                                        & Random                                                                            & 0.511& 0.555& 0.492                                                  \\ \hline

ViViT& $r = 0.25 $& Empty                                                        & Prog.-aware                                                                       & 0.482& 0.530& 0.481\\ \hline

ViViT& $r = 0.5 $& Empty                                                        & Prog.-aware                                                                       & 0.497& 0.512& 0.622\\ \hline

ViViT& $r = 0.75 $                                                 & Empty                                                        & Prog.-aware                                                                       & 0.499 & 0.553     & 0.628\\ \hline  

ViViT& $r_{mask}=$ 0.25                                              
& Base                                                         & Random                                                                            & 0.521& 0.521                                                                     & 0.519                                                  \\ \hline  

ViViT& $r_{mask}=$ 0.5                                               
& Base                                                         & Random                                                                            & 0.511                                                                    & 0.555                                                                     & 0.492                                                  \\ \hline  

ViViT& $r_{mask}=$ 0.75                                              
& Base                                                         & Random                                                                            & 0.488                                                                    & 0.529                                                                     & 0.614                                                  \\ \hline  

ViViT& $r = 0.25 $
& Base                                                         & Prog.-aware                                                                       & 0.499& 0.567& 0.611\\ \hline  

ViViT& $r = 0.5 $
& Base                                                         & Prog.-aware                                                                       & 0.480& 0.541& 0.631\\ \hline  

ViViT& $r = 0.75 $                                                 & Base                                                         & Prog.-aware                                                                       & 0.495 & 0.589& 0.633\\ \hline  

Time-ViViT  & $r_{mask}=$ 0.25                                              
& Time-aware                                                   & Random                                                                            & 0.488                                                                    & 0.546                                                                     & 0.690                                                  \\ \hline  

Time-ViViT  & $r_{mask}=$ 0.5                                               
& Time-aware                                                   & Random                                                                            & 0.536                                                                    & 0.530                                                                     & 0.728                                                  \\ \hline  

Time-ViViT  & $r_{mask}=$ 0.75                                              
& Time-aware                                                   & Random                                                                            & 0.485                                                                    & 0.541                                                                     & 0.748                                                  \\ \hline  

Time-ViViT  & $r = 0.25 $
& Time-aware                                                   & Prog.-aware                                                                       & 0.488& 0.546& 0.690\\ \hline 

Time-ViViT  & $r = 0.5 $
& Time-aware                                                   & Prog.-aware                                                                       & 0.565& 0.571& 0.723\\ \hline

Time-ViViT  & $r = 0.75 $                                                 & Time-aware                                                   & Prog.-aware                                                                       & 0.560                                                           & 0.584& 0.797                                         \\ \hline

\end{tabular}%
}
\end{table}

\subsubsection{Weight initialization importance}

One of the motivations of our paper lies in the fact that pre-training initialization of ViTeffectiveness~\cite{dosovitskiy2021image} has been demonstrated only when it is trained on large datasets since Transformers lack certain inductive biases of convolutional networks~\cite{dosovitskiy2021image}. Consequently, it is challenging to train large models from the beginning to a high level of accuracy. By showing that they are essential for the model to achieve a high accuracy. Illustrate the significance of the three pretrained weights element of our model by demonstrating that they are indispensable for the model to attain a high precision. The three main weight components are layer embedding, time-aware embedding, and the encoder. We have tried different combinations of initialization weights, with random weights and weights obtained with the best model. According to Tab.~\ref{table:results_comparaison_weight_init}, the initialization of the weights plays a crucial role in order to obtain better performance. While all the weights do share relevant information, we observe that the temporal embedding and encoder weights are more discriminating for the prediction of the development of Severe+ RD. Conversely, the embedding layer seems to play a minor role across the different tasks.

\begin{table}
\centering
\caption{Comparison of approaches with different weight initialization. With \cmark ~and \xmark~ denoting with and without the weight initialization, respectively.}
\label{table:results_comparaison_weight_init}

\resizebox{\columnwidth}{!}{%

\begin{tabular}{|l|c|c|l|l|l|l|} 
\hline
Method & Embeddings layer & \begin{tabular}[c]{@{}c@{}}Temporal\\ embedding\end{tabular} & \multicolumn{1}{|c|}{\begin{tabular}[c]{@{}c@{}}Encoder\\weights\end{tabular}} & \multicolumn{1}{|c|}{\begin{tabular}[c]{@{}c@{}}AUC\\ Mild+\end{tabular}} & \multicolumn{1}{|c|}{\begin{tabular}[c]{@{}c@{}}AUC\\ Moder+\end{tabular}} & \begin{tabular}[c]{@{}l@{}}AUC\\ Severe+\end{tabular}  \\ 
\hline
Time-ViViT  & \xmark               & \xmark                                                           & \xmark                                                                            & 0.500                                                                    & 0.528                                                                     & 0.454                                                  \\ 
\hline
Time-ViViT  & \cmark                & \xmark                                                           & \xmark                                                                            & 0.480& 0.541& 0.631\\ \hline 
 Time-ViViT  & \xmark& \cmark& \xmark                                                                            & 0.527 & 0.519     & 0.729   \\ \hline 
 Time-ViViT  & \xmark& \xmark                                                           & \cmark& 0.530& 0.546&0.710\\ 
\hline
Time-ViViT  & \cmark                & \cmark                                                            & \xmark                                                                            & 0.498& 0.484& 0.709\\ \hline 
 Time-ViViT  & \cmark                & \xmark& \cmark& 0.566& 0.590&0.744\\ \hline 
 Time-ViViT  & \xmark& \cmark                                                            & \cmark& 0.563& 0.586&0.790\\\hline 
\hline
Time-ViViT  & \cmark                & \cmark                                                           & \cmark                                                                             & 0.560& 0.584& 0.797\\
\hline
\end{tabular}%
}
\end{table}

\subsubsection{Progressive masking-aware strategies}

Among the hyperparameters that change the nature of the masking, there is the value ratio of $r$. We tried three values of $r$. The effect of the change in value can be seen in Fig.~\ref{fig:overview}. The more $r$ increases, the larger the central region is masked. An increasing random noise is added on top of the masking depending on the severity grade of the image; a higher severity grade results in more regions masked in the periphery. Based on the Tab.~\ref{table:results_comparaison_hp_masking} result in terms of AUC indicates that the best value of $r$ is 0.75. Moreover, we also observe that the more we increase the value of the masking parameter for both random and progression-aware schemes, the better the results.

\subsubsection{Importance of the component of our proposed method}

\begin{table}[H]
\centering
\caption{Comparison of approaches in terms of the different AUC scores for the predict next task, on the test set regarding the importance of each component of our proposed method.}
\label{table:results_comparaison_component_importance}
\begin{tabular}{|l|c|l|l|l|l|} 
\hline
Method     & \begin{tabular}[c]{@{}c@{}}Temporal\\ embedding\end{tabular} & \multicolumn{1}{c|}{\begin{tabular}[c]{@{}c@{}}Masking\\ strategies\end{tabular}} & \multicolumn{1}{c|}{\begin{tabular}[c]{@{}c@{}}AUC\\ Mild+\end{tabular}} & \multicolumn{1}{c|}{\begin{tabular}[c]{@{}c@{}}AUC\\ Moder+\end{tabular}} & \begin{tabular}[c]{@{}l@{}}AUC\\ Severe+\end{tabular}  \\ 
\hline
ViViT      & No                                                           & Random                                                                            & 0.511                                                                    & 0.555                                                                     & 0.492                                                  \\ 
\hline
ViViT      & No                                                           & Prog.-aware                                                                       & 0.499                                                                    & 0.553                                                                     & 0.628                                                  \\ 
\hline
ViViT      & Base                                                         & Random                                                                            & 0.488                                                                    & 0.529                                                                     & 0.614                                                  \\ 
\hline
ViViT      & Base                                                         & Prog.-aware                                                                       & 0.495                                                                    & 0.589                                                                     & 0.633                                                  \\ 
\hline
Time-ViViT & Time-aware                                                   & Random                                                                            & 0.485                                                                    & 0.541                                                                     & 0.748                                                  \\ 
\hline
Time-ViViT & Time-aware                                                   & Prog.-aware                                                                       & 0.560                                                                    & 0.584                                                                     & 0.797                                                  \\
\hline
\end{tabular}
\end{table}

The table \ref{table:results_comparaison_component_importance} presents the performance of two variants of ViViT, a vision transformer architecture, for the task of classifying medical images into three severity levels three years later: mild, moderate, and severe. The two variants differ in their temporal embedding and masking strategies. The temporal embedding encodes the temporal information in the image sequence, while the masking strategy determines which frames to use for training and evaluation. The results show that both temporal embedding and masking strategies have a significant impact on the model performance. The Time-ViViT variant, which uses a time-aware temporal embedding and a progressive-aware masking strategy, consistently outperforms the ViViT variant, which does not use temporal embedding or a masking strategy. This suggests that explicitly modeling temporal information and dynamically masking zones that are correlated with the progression of DR are crucial for improving the performance of ViViT for this task. More specifically, the Time-ViViT variant achieves an AUC of 0.560 for mild+, 0.584 for moder+, and 0.797 for severe+, which is significantly higher than the AUCs of 0.511, 0.555, and 0.492 for the ViViT variant, respectively. This indicates that Time-ViViT is better at distinguishing between mild and moderate cases, and it is also significantly better at identifying severe cases. The improved performance of Time-ViViT can be attributed to its ability to capture temporal dependencies in the image sequence. The time-aware temporal embedding allows the model to learn the temporal relationships between frames, while the progressive-aware masking strategy ensures that the model pays attention to the most relevant frames for the task. Overall, the results demonstrate that Time-ViViT is a more effective architecture for classifying medical images into severity levels than ViViT. The use of temporal embedding and masking strategies can significantly improve the model performance, especially for identifying severe cases. These findings have implications for the development of automated systems for medical image analysis.

\subsubsection{Masking strategies} 

We tried two other masking strategies. The first is the proposed masking strategy, but instead of masking based on the value of the image label, we randomly assigned a severity grade to an image and masked it according to Fig.~\ref{fig:proposed_method}. In order to demonstrate whether using at random label, the masking strategy was better than our masked strategy guided by the severity grade of DR progression.

\begin{table}[H]
\centering
\caption{Comparison of approaches in terms of masking strategies during the pretraining phase for the
different AUC scores for the predict next task, on the test set. Results in bold show the best results.}
\label{table:results_comparaison_masking}
\begin{tabular}{|l|l|l|l|l|} 
\hline
Method & \multicolumn{1}{|c|}{\begin{tabular}[c]{@{}c@{}}Masking\\ strategies\end{tabular}} & \multicolumn{1}{|c|}{\begin{tabular}[c]{@{}c@{}}AUC\\ Mild+\end{tabular}} & \multicolumn{1}{|c|}{\begin{tabular}[c]{@{}c@{}}AUC\\ Moder+\end{tabular}} & \begin{tabular}[c]{@{}l@{}}AUC\\ Severe+\end{tabular}  \\ 
\hline
Time-ViViT  & Random                                                                            & 0.485                                                                    & 0.541                                                                     & 0.748                                                  \\ 
\hline
Time-ViViT  & Visit                                                                            & 0.530& 0.546& 0.710\\ 
\hline
Time-ViViT  & Prog-aware random                                                                           & 0.558& 0.571& 0.760\\ 
\hline
Time-ViViT  & Prog.-aware                                                                       & \textbf{0.560}                                      & \textbf{0.584}                                       & \textbf{0.797}                                         \\ \hline
\end{tabular}
\end{table}

The second was to mask an entire frame at random using a uniform distribution between the set of available images in the sequence in the time dimension. Motivated by the fact that removing a visit would force the model to take advantage of the remaining visit and capture the overall progression of the patient's trajectory with the global self-attention layers across all tokens. Conformable to Tab.~\ref{table:results_comparaison_masking}, the visit masking strategies perform worse than the baseline random masking for the different tasks. Surprisingly, the progression-aware masking with a random choice performs better than the baseline. This indicates that the form of the masking strategies is more relevant for predicting the DR's development than a random form. However, it performs less than the standard progression-aware masking. This supports our argument that masking strategies aligned with disease progression characteristics are a more informative masking strategy for the MAE model for predicting the development of diabetic retinopathy \cite{WAHEED2023101206,STINO2023}.

\section{Discussion}

%%% part in the discussion

In this study, we tackle the task of predicting the progression of diabetic retinopathy, and we observed that the models based on using time perform better than those not time-aware. The use of a model not based on time to solve a task that tackles irregularly sampled time series is not optimal like it was mentioned in previous studies \cite{Li2020LearningFI,Sun2020ARO,Rubanova}, which is also reflected by our results. One interesting aspect that emerged from the analysis is that the masking strategies for the MAE are clearly important. Adding medical knowledge to the masking strategy brings out a better representation according to the presented results. 

However, we note that our masking strategy performs the best compared to the different random masking strategies, providing a better representation for the late stage according to the \textbf{AUC Moderate+} and \textbf{AUC Severe+}. We suspect that it is due to the design of our mask that forces the model to reconstruct highly discriminating zones, resulting in an enhanced representation for predicting the progression of DR. This finding is notably observed for the experiments where an empty embedding was compared for the different masking strategies. This result is aligned with the following study \cite{mao2023medical}, which performs masking auto encoding pretraining and observed that the masking area is the crucial element in the success of getting a good feature extractor that performs well on the downstream task. 

Our findings support \cite{he2021masked,feichtenhofer2022masked} hypothesis that masking a relatively large region is more effective in creating a more robust feature extractor.

Yet some limitations should be pointed out. There is a large number of hyper-parameters in the proposed masking strategy, and the creation of masks is less efficient in terms of computation power than classical random masking. In addition, we also think that our proposed masking strategy could shine better using larger fields of view images based either on a mosaic of CFP or ultra-widefield color fundus photographs (UWF-CFP) in order to fully take advantage of our contributions. As an extension and to overcome the mentioned limitation, we would like to learn an optimal mask using a semi-supervised technique through attention maps with Transformer-based method \cite{Sun2021LesionAwareTF,li2022semmae} while keeping the idea of a progressive mask according to the increase of DR severity applied to UWF-CFP or CFP mosaics. 

It should be noted that neither the baselines nor any configuration of the Time-ViViT can produce satisfactory results in terms of AUC, for the prediction of the development of DR in the early stages. These different results can be partly explained by the fact that the images we used are relatively small compared to their original size and that the most critical characteristic indicates a small lesion or that the image alone is insufficient to establish the future progression of diabetic retinopathy, especially for early grades. 

It should also be noted that the task of predicting the future development of diabetic retinopathy is challenging due to the following two problems. The first is that a defined task is challenging since deep learning models are not well-equipped to capture temporal dependencies, which are the relationships between events that occur at different times. This can make it difficult to use them to predict disease progression that occurs in the future, and the second one is the fact most tasks that are defined to predict the disease progression within a certain period of time often remove many patients that do not respect the selection criteria. An alternative could be to use the time to condition the future prediction point as it was performed in \cite{LMT}. These approaches allow the model to train without restriction, which was not made explicit in the article.  

Our method inherits the quadratic complexity of self-attention \cite{keles2022computational} with respect to the number of tokens of a Transformer \cite{vaswani2017attention}. This complexity for longitudinal image data, as the number of tokens grows in a linear fashion with the number of input frames, encourages the creation of more efficient architectures. In this regard, researchers are working to develop new techniques to reduce the memory usage of Transformer-based methods. These techniques include developing new model architectures that are more efficient in terms of memory usage, developing new sparse attention techniques \cite{Tay2020SparseSA,liu2021transformer} that can reduce the number of attention weights that need to be computed, and developing new training and inference techniques that can be used with transformer-based models on devices with limited memory resources \cite{dao2022flashattention}.

Our results in Tab.~\ref{table:results_comparaison} and Tab.~\ref{table:results_comparaison_component_importance} and ablation study have demonstrated the complexity of such tasks. To alleviate this problem, one possible way would be to inject even more context into our approach by means of multimodal data based on other retina modalities \cite{li_3-d_2023,daho2024discover} or on non-imaging data \cite{Bora2021,daiDeepLearningSystem2024}. Typical approaches related to disease progression in DR often rely on clinical data, specifically HbA1c \cite{Bora2021,daiDeepLearningSystem2024} . Adding this information could further improve the predictive capacity of our framework, especially for early stages. 

\section{Conclusion}

In this paper, we proposed to apply MAE to longitudinal data with two simple but effective extensions. The results demonstrate the importance of injecting time in the positional embedding to increase the predictive ability of the Transformer. Moreover, the proposed masking strategy exhibits a more discriminate nature to detect late stages of DR compared to standard random masking, which is in accordance with clinical studies \cite{WAHEED2023101206}. In conclusion, our research has laid the groundwork for using Transformers for longitudinal and disease progression tasks. The outcomes are promising, and further exploration could assist medical professionals in selecting the most suitable DR screening intervals. Our framework could be extended to other retinal problems that involve longitudinal studies. It would also be beneficial to examine the effect of our masking approach on other retinal diseases, as many of these conditions are classified by the amount, size, and kind of lesions.

% \appendix

% \section{Sample Appendix Section}
% \label{sec:sample:appendix}
% Lorem ipsum dolor sit amet, consectetur adipiscing elit, sed do eiusmod tempor section \ref{sec:sample1} incididunt ut labore et dolore magna aliqua. Ut enim ad minim veniam, quis nostrud exercitation ullamco laboris nisi ut aliquip ex ea commodo consequat. Duis aute irure dolor in reprehenderit in voluptate velit esse cillum dolore eu fugiat nulla pariatur. Excepteur sint occaecat cupidatat non proident, sunt in culpa qui officia deserunt mollit anim id est laborum.

%% If you have bibdatabase file and want bibtex to generate the
%% bibitems, please use
%%
\bibliographystyle{elsarticle-num-names} 
\bibliography{cas-refs}

%% else use the following coding to input the bibitems directly in the
%% TeX file.

% \begin{thebibliography}{00}

% %% \bibitem[Author(year)]{label}
% %% Text of bibliographic item

% \bibitem[ ()]{}

% \end{thebibliography}
\end{document}